\documentclass[conference]{IEEEtran}

\ifCLASSINFOpdf
  
\else
 
\fi

\usepackage{graphicx}
\usepackage{subfig}
\usepackage{breqn}
\usepackage{balance}
\usepackage[ruled,vlined,linesnumbered]{algorithm2e}
\usepackage{enumitem,kantlipsum}

\begin{document}

\title{TAnoGAN: Time Series Anomaly Detection with Generative Adversarial Networks}


\author{\IEEEauthorblockN{Md Abul Bashar}
\IEEEauthorblockA{School of Computer Science\\
Centre for Data Science\\
Queensland University of Technology\\
Brisbane, Queensland 4000, Australia\\
Email: m1.bashar@qut.edu.au}
\and
\IEEEauthorblockN{Richi Nayak}
\IEEEauthorblockA{School of Computer Science\\
Centre for Data Science\\
Queensland University of Technology\\
Brisbane, Queensland 4000, Australia\\
Email: r.nayak@qut.edu.au}}

    \IEEEoverridecommandlockouts
    \IEEEpubid{\makebox[\columnwidth]
    {978-1-7281-2547-3/20/\$31.00~\copyright2020 IEEE \hfill} 
    \hspace{\columnsep}\makebox[\columnwidth]{ }}
    \IEEEpubidadjcol

\maketitle
\begin{abstract}
Anomaly detection in time series data is a significant problem faced in many application areas such as manufacturing, medical imaging and cyber-security. Recently, Generative Adversarial Networks (GAN) have gained attention for generation and anomaly detection in image domain. In this paper, we propose a novel GAN-based unsupervised method called TAnoGan for detecting anomalies in time series when a small number of data points are available. We evaluate TAnoGan with 46 real-world time series datasets that cover a variety of domains. Extensive experimental results show that TAnoGan performs better than traditional and neural network models.
\end{abstract}


\IEEEpeerreviewmaketitle

\vspace{-2mm}
\section{Introduction}
\vspace{-2mm}
The ubiquitous use of networked sensors and actuators in places like smart buildings, factories, power plants and data centres as well as the emergence of the Internet of Things (IoT) have resulted in generating substantial amounts of time series data. These data can be used to continuously monitor the working conditions of these environments to detect anomalies. Anomalies are system behaviour patterns in time steps that do not conform to a well-defined notion of normal behaviour \cite{Chandola2008ComparativeData}. Anomaly detection provides the opportunity to take actions to investigate and resolve the underlying issue before it causes disasters. Closely monitoring these environments is necessary as many of them are deployed for mission-critical tasks.

Anomaly detection in time series data is commonly approached as an unsupervised machine learning task due to the lack of label information \cite{Li2019MAD-GAN:Networks,Schlegl2017UnsupervisedDiscovery,DiMattia2019ADetection}. Most existing unsupervised anomaly detection methods (e.g. \cite{Li2014ATransform,wold1985encyclopedia}) use linear projection and transformation \cite{Li2019MAD-GAN:Networks}. However, such methods cannot handle non-linear interactions in time series data. Another family of methods compares the system state value at current time step and the predicted normal range to detect anomalies \cite{DiMattia2019ADetection}. These methods largely fail as most of the systems are highly dynamic in nature and it is difficult to define a normal range of measurements.

The Generative Adversarial Networks (GAN) framework is increasingly becoming popular for learning generative models through adversarial training \cite{GoodfellowIanandPouget-AbadieJeanandMirzaMehdiandXuBingandWarde-FarleyDavidandOzairSherjilandCourvilleAaronandBengio2014GenerativeNets}. A GAN model can successfully generate realistic images \cite{GoodfellowIanandPouget-AbadieJeanandMirzaMehdiandXuBingandWarde-FarleyDavidandOzairSherjilandCourvilleAaronandBengio2014GenerativeNets,DentonDeepNetworks,DonahueADVERSARIALLEARNING} and synthetic data \cite{Esteban2017Real-valuedGANs,Mogren2016C-RNN-GAN:Training}. It has been used to model a complex and high-dimensional general (i.e., normal) distribution of real-world time series data \cite{Esteban2017Real-valuedGANs}. 
Recently a GAN framework coupled with the mapping of data to latent space has been explored for anomaly detection \cite{Schlegl2017UnsupervisedDiscovery,Li2019MAD-GAN:Networks}. While GAN has been extensively investigated in image domain for generation and anomaly detection, only a few works (e.g. \cite{Esteban2017Real-valuedGANs,Li2019MAD-GAN:Networks}) have explored the potential of GAN in time series domain.

In this paper, we propose a novel method, Time series Anomaly detection with GAN (TAnoGan)\footnote{The code is available at https://github.com/mdabashar/TAnoGAN}, for unsupervised anomaly detection in time series data when a small number of data points are available. Detecting anomalies in time series using GAN requires modelling the normal behaviour of time series data using the adversarial training process and then detecting anomalies using an anomaly score that indicates how much the data points have deviated from the normal behaviour \cite{Schlegl2017UnsupervisedDiscovery,Zenati2018EfficientDetection,Li2019MAD-GAN:Networks}. For learning the anomaly score, we first map the real time series data space to a latent space and then reconstruct the data from latent space. The anomaly score is estimated as the loss between the real data and reconstructed (fake) data. 

The architecture of TAnoGan is designed to detect anomalies in datasets with a small number of data points. We evaluated TAnoGan with the diverse 46 time series datasets in the Numenta Anomaly Benchmark (NAB) time series data collection that cover a variety of domains \cite{Lavin2017ThePaper,Lavin2015EvaluatingBenchmark}, each with a small number of data points ranging from one thousand to 22 thousands only. We evaluated TAnoGan against the neural network model Long Short-Term Memory (LSTM).  TAnoGan provides significantly improved results. These results ascertain that LSTM-based GAN improves over LSTM by utilising adversarial training.

The main contributions of this paper are: (1) We propose a GAN-based method called TAnoGan to detect anomalies in time series datasets when a small number of data points are available. In general, neural network-based models require setting a large number of parameters. Therefore, these models need to be trained on huge number of data points. (2) We evaluate TAnoGan on an extensive and diverse range of 46 time series datasets from the NAB collection that cover a variety of domains. (3) We show that LSTM-based GAN improves over LSTM by utilising adversarial training in time series data. (4) We show that TAnoGan performs better than traditional and neural network models.

The rest of this paper is organised as follows. Section \ref{sec:related_work} gives a summary of related work. Section \ref{sec:TAnoGAN} describes the proposed TAnoGan method. Empirical evaluation of the proposed model is given in Section \ref{sec:evaluation}. The paper concludes in Section \ref{sec:conclusions}. 

\vspace{-4mm}
\section{Related Work}
\label{sec:related_work}
\vspace{-.1cm}
Principal Component Analysis (PCA) \cite{Li2014ATransform} and Partial Least Squares (PLS) \cite{wold1985encyclopedia} are two popular linear model-based unsupervised anomaly detection methods. However, these models assume Gaussian distribution in the data and are only effective for highly correlated data \cite{Dai2013FromDiagnosis}.
K-Nearest Neighbor (KNN) is a popular distance-based method \cite{Angiulli2002FastSpaces}. However, distance-based methods assume prior knowledge about anomaly duration and the number of anomalies. 
Angle-Based Outlier Detection (ABOD) \cite{Kriegel2008Angle-basedData} and Feature Bagging (FB) \cite{Lazarevic2005FeatureDetection} are density estimation-based probabilistic models that outperform distance-based methods. However, these methods do not work well for time series data as they do not consider the temporal correlations \cite{Li2019MAD-GAN:Networks}.

Gaussian Mixture Model (GMM) \cite{bishop2006linear,Reynolds2009GaussianModels}, Isolation Forest (IsoF) \cite{Liu2008IsolationForest}, One Class Support Vector Machine (OCSvm) \cite{ScholkopfSupportDetection} are commonly used in practical anomaly detection applications. Our experiments in this paper show that GMM perform well but IsoF and OCSvm give poor performance. A short description of these models are given in Section \ref{sec:baselines}.  

Recently, deep learning-based unsupervised anomaly detection methods such as Auto-Encoder \cite{Zhou2017AnomalyAutoencoders}, Encoder-Decoder \cite{Habler2018UsingMessages} and LSTM \cite{Malhotra2015LongSeries,hochreiter1997long} have gained popularity due to their promising performance. Latest development is anomaly detection models built on the GAN framework. The majority of GAN based methods have been applied in the image domain. A GAN-based method allows to learn generative models that can generate realistic images \cite{GoodfellowIanandPouget-AbadieJeanandMirzaMehdiandXuBingandWarde-FarleyDavidandOzairSherjilandCourvilleAaronandBengio2014GenerativeNets,DentonDeepNetworks,DonahueADVERSARIALLEARNING}. 

\cite{Schlegl2017UnsupervisedDiscovery} used a GAN based unsupervised learning model namely AnoGan to identify anomalies in medical imaging data. AnoGan uses an adversarial network to learn normal anatomical variability. It then uses an anomaly scoring scheme by mapping images from image space to a latent space and reconstructing the images. The loss in a reconstructed image is used to estimate anomaly score. The anomaly score of an image indicates its deviation from the general data distribution and can identify anomaly images. This model uses Convolutional Neural Network (CNN) in its generator and discriminator to effectively identify anomalies in images. However, the generator in the model cannot be effectively used in time series datasets as it does not include any mechanism to handle time sequence. On the other hand, \cite{Esteban2017Real-valuedGANs} used LSTM in both generator and discriminator of a GAN-based model to generate realistic time series data in medical domain. However, \cite{Esteban2017Real-valuedGANs} was not designed to detect anomalies. 

In this paper, we propose a GAN-based model, named TAnoGan, to detect anomalies in time series datasets when a small number of data points are available. We propose using LSTM as the generator and discriminator model to handle the time series data. We evaluate the proposed LSTM based GAN model against neural network-based models such as Auto-Encoder \cite{Zhou2017AnomalyAutoencoders}, Encoder-Decoder \cite{Habler2018UsingMessages} and LSTM \cite{Malhotra2015LongSeries,hochreiter1997long} to see the improvement due to adversarial training compared with traditional training. The GAN model in \cite{Li2019MAD-GAN:Networks} was found effective for datasets with a large number of data points, e.g., two large time series datasets each with about one million data points. Usually, neural network-based models require huge number of data points for setting their large number of parameters \cite{bashar2018cnn}. Different from \cite{Li2019MAD-GAN:Networks}, we use a different and effective architecture to detect anomalies in small datasets. Investigating the feasibility of using GAN for detecting anomalies in time series datasets with small number of data points will open the potential of using GAN in many application domains. We investigate the effectiveness of GAN in anomaly detection on a variety of datasets to assess how it generalises in this task. 



\section{Generative Adversarial Representation Learning to Identify Time Series Anomalies}
\label{sec:TAnoGAN}

\begin{figure*}[htb!]
    \centering
    \subfloat[Representing Normal Time Series with Generator $G$]{{\includegraphics[width=9cm]{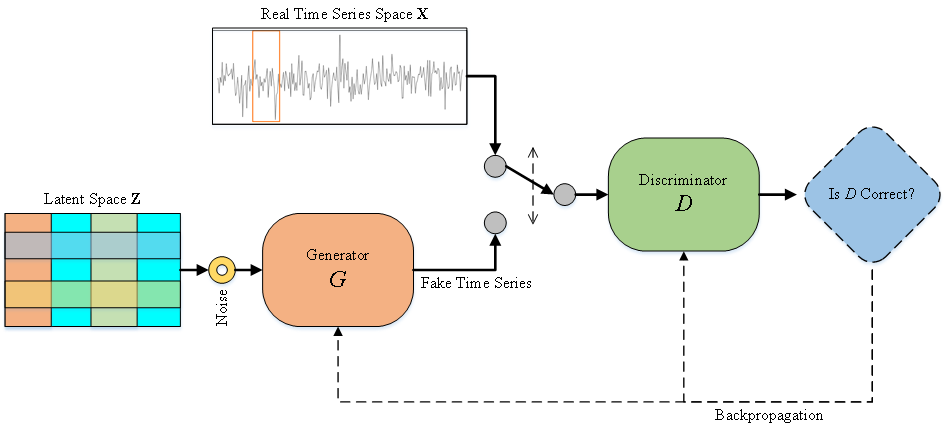}}\label{TAnoGAN_StepOne}}
    \quad
    \subfloat[Mapping Real-Data to the Latent Space]{{\includegraphics[width=7cm]{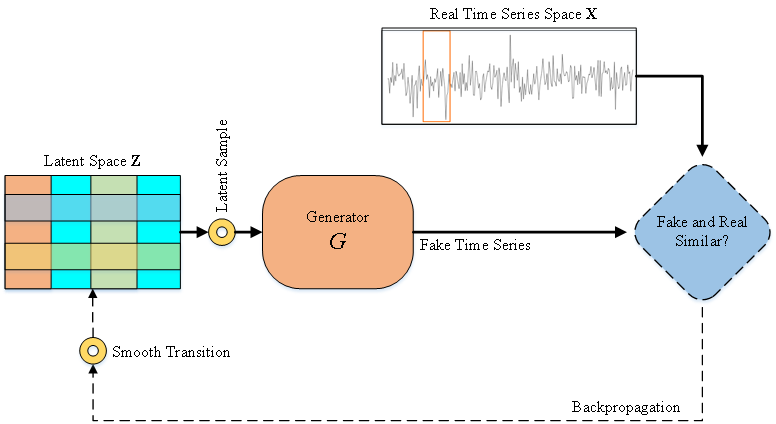}}\label{TAnoGAN_StepTwo}}
    \caption{TAnoGan: Time Series Anomaly Detection with Generative Adversarial Networks}
    \label{fig:TAnoGAN}
    \vspace{-.6cm}
\end{figure*}

The main idea of unsupervised anomaly detection in time series data is to identify whether data observations conform to the normal data distributions over the time. The non-conforming observations are identified as anomalies \cite{Chalapathy2019DeepSurvey,Kwon2019ADetection}. Figures \ref{TAnoGAN_StepOne} and \ref{TAnoGAN_StepTwo} show the two sup-processes in the proposed TAnoGan method. A pseudocode for the proposed TAnoGan model is given in Algorithm 1. In the first sub-process (Figure \ref{TAnoGAN_StepOne}), we learn a model representing normal time series variability based on Generative Adversarial Networks (GAN) \cite{Schlegl2017UnsupervisedDiscovery,Li2019MAD-GAN:Networks}. After this sub-process, the generator can generate realistic (fake) time series sequences from a latent space. In the second sub-process (Figure \ref{TAnoGAN_StepTwo}), we map real time series sequences to a latent space and reconstruct the sequences from the latent space. The reconstruction loss is used to identify anomalies.

\vspace{-.1cm}
\subsection{Learning General Data Distribution}
\vspace{-.1cm}
The aim of this process is to learn the general data distribution of a given dataset through an adversarial training. 
This process simultaneously trains a Generator $G$ that generates fake time series data and a discriminator $D$ that learns to distinguish between the generated fake data and the real data. To handle the time-series data, we propose to use a Long Short-Term Memory (LSTM) as a generator model and another LSTM as a discriminator model. 

To address the small number of data points problem, we have investigated a variety of architectures for generator and discriminator. We observed that when the dataset is small, a large discriminator easily overfits the data and a shallow generator cannot generate data realistic enough to beat the discriminator. We propose to use a simple shallow discriminator and a medium depth generator when the dataset is small. We observed that starting from a small number of hidden units and progressively increasing the number of hidden units in each successive layer is effective for training the generator on small datasets. Driven by the experimental results, we use the LSTM in generator model $G$ with three stacked layers with 32, 64 and 128 hidden units. We use a single layer LSTM in discriminator model $D$ with 100 hidden units.

The input to the generator is a noise vector $\mathbf{z}$ randomly selected from the latent space $Z$. A time series of original data is divided into small sequences with a sliding window $s_w$ before sending it to discrimination $D$. $G$ generates similar (fake) small sequences. This process resembles as if the Generator $G$ has the Discriminator $D$ as an adversary \cite{GoodfellowIanandPouget-AbadieJeanandMirzaMehdiandXuBingandWarde-FarleyDavidandOzairSherjilandCourvilleAaronandBengio2014GenerativeNets}. $G$ needs to learn how to generate data in such a way that $D$ cannot distinguish it as fake anymore. The competitiveness between $G$ and $D$ improves their knowledge and continues until $G$ succeeds in generating realistic time series data by learning the general data distribution accurately.

$G(z, \theta_1)$ function models the Generator that maps input noise vectors $\mathbf{z} \in \mathbf{Z}$ to the desired data space $\mathbf{x} \in \mathbf{X}$ (i.e. time series small sequences). $D(x, \theta_2)$ function models the Discriminator that outputs the probability that the data is real. Here $\theta_1$ and $\theta_2$ are parameters of the models. The loss function of this network maximises the function $D(\mathbf{x})$ and minimises $D(G(\mathbf{z}))$. 
After enough iteration of training, $G$ and $D$ will reach a point at which they cannot improve anymore. At this point, the $G$ generates realistic time series data, and $D$ is unable to differentiate between fake and real data.
 
Both $G$ and $D$ attempt to optimise the competitive loss functions in the training period. Hence, they can be considered as two agents playing a minimax game with a value function $V(G,D)$. $G$ tries to maximise the probability of $G(\mathbf{z})$ being recognised as real, while $D$ tries to minimise the same value. The value function $V(G,D)$ is defined as follow \cite{GoodfellowIanandPouget-AbadieJeanandMirzaMehdiandXuBingandWarde-FarleyDavidandOzairSherjilandCourvilleAaronandBengio2014GenerativeNets}.  
\begin{dmath}
\min_G \max_D V(D,G) = 
E_{\mathbf{x}\sim p_{data}(\mathbf{x})}[log D(\mathbf{x})] + E_{\mathbf{z}\sim p_z(\mathbf{z})}[log(1-D(G(\mathbf{z})))],
\end{dmath} 
where $E_{\mathbf{x}}$ is the expected value of $\mathbf{x}$ and $E_{\mathbf{z}}$ is the expected value of $\mathbf{z}$.
 
Stochastic gradient descent (SGD) has proven to be successful for finding an optimised solution in multiple fields \cite{Chalapathy2019DeepSurvey,Kwon2019ADetection,GoodfellowIanandPouget-AbadieJeanandMirzaMehdiandXuBingandWarde-FarleyDavidandOzairSherjilandCourvilleAaronandBengio2014GenerativeNets}. Therefore, we use SGD to train the GAN network.
Once this adversarial training is completed, the real time series small sequences $\mathbf{x} \in \mathbf{X}$ are mapped to the latent Space $\mathbf{z} \in \mathbf{Z}$ for anomaly detection. 

\begin{algorithm}
\scriptsize
\caption{Algorithm for TAnoGan}
\SetKwInput{KwInput}{Input} 
\SetKwInput{KwOutput}{Output} 

\DontPrintSemicolon
  \KwInput{A list of small sequences $\mathbf{X}$.}
  \KwOutput{A list of anomaly scores $A$.}

  \SetKwFunction{FMain}{Main}
  \SetKwFunction{FAdvTrn}{adversarialTrain}
  \SetKwFunction{FMap}{anomalyScore}
 
    \SetKwProg{Fn}{Function}{:}{}
    \Fn{\FAdvTrn{$\mathbf{X}$}}{
        \For{number\_of\_epochs}{
            Sample $m$ noise vectors $\{\mathbf{z}_1, \dots \mathbf{z}_m\}$ from the noise prior $p_g(\mathbf{z})$.\\
            Generate $m$ fake-data vectors $\{G(\mathbf{z}_1), \dots, G(\mathbf{z}_m)\}$ from the $m$ noise vectors.\\
            Sample $m$ real-data vectors $\{ \mathbf{x}_1, \dots, \mathbf{x}_m \}$ from the data generating distribution $p_{data}(\mathbf{x})$.\\
            Train $D$ on the fake-data vectors and real-data vectors.\\
            Sample another $m$ noise vectors $\{\mathbf{z}_1, \dots \mathbf{z}_m\}$ to from the noise prior $p_g(\mathbf{z})$.\\
            Train $G$ on the second set of noise vectors. 
        }
        \KwRet $G$, $D$
    }
  \;
    \SetKwProg{Fn}{Function}{:}{}
    \Fn{\FMap{$\mathbf{X}$, $G$, $D$}}{
        \For{$i$ in 1 to $m$}{
            Sample a noise vector $\mathbf{z}^i$ from the noise prior $p_g(\mathbf{z}^i)$.\\
            \For{$\lambda$ in 1 to $\Lambda$}{
                Generate a fake-data vector $G(\mathbf{z}^i)$ from the noise vector $\mathbf{z}^i$.\\
                Calculate $\mathcal{L}(G(\mathbf{z}^i))$ for $\mathbf{x}^i$ utilising $G$ and $D$, and update $\mathbf{z}^i$ using gradient descent.
            }
            $A(\mathbf{x}^i)$ = $\mathcal{L}(G(\mathbf{z}^i))$
        }
        \KwRet $A$
    }
    \;
    \SetKwProg{Fn}{Function}{:}{\KwRet}
    \Fn{\FMain{$\mathbf{X}$}}{
        $G$, $D$ = adversarialTrain($\mathbf{X}$)\\
        A = anomalyScore($\mathbf{X}$, $G$, $D$)
    }
\end{algorithm}
\subsection{Mapping Real-Data to the Latent Space}
In the adversarial training, the generator learns the mapping $G: \mathbf{Z} \rightarrow \mathbf{X}$ such that $G(\mathbf{z}) \in \mathbf{X}$, i.e. from latent space representations $\mathbf{z}$ to realistic (normal) time series small sequences $\mathbf{x}$.
To detect anomaly, first we need to map real time series small sequences $\mathbf{x} \in \mathbf{X}$ to the latent space $\mathbf{z} \in \mathbf{Z}$ to see how closely the corresponding latent space generates real time series small sequences. However, GAN does not have inverse mapping $G^{-1}: \mathbf{X} \rightarrow \mathbf{Z}$ such that $G^{-1}(\mathbf{x}) \in \mathbf{Z}$.

Given a real time series small sequence $\mathbf{x} \in \mathbf{X}$, we need to find $\mathbf{z} \in \mathbf{Z}$ corresponding to a small sequence $G(\mathbf{z})$ that is most similar to real small sequence $\mathbf{x}$. The degree of similarity between $\mathbf{x}$ and $G(\mathbf{z})$ depends on the extent $\mathbf{x}$ follows the data distribution $p_g$ that was used for training the Generator $G$.

Figure \ref{TAnoGAN_StepTwo} shows the 
process of mapping real time series small sequences to the latent space. To find the best $\mathbf{z}$ for a given $\mathbf{x}$, this process starts with randomly sampling $\mathbf{z}_1 \in \mathbf{Z}$ and feeding it into the trained generator $G$ to get a fake small sequence $G(\mathbf{z}_1)$. Based on the fake small sequence $G(\mathbf{z}_1)$, we define a loss function (described below) $\mathcal{L}$ that provides gradients to update the parameters of $\mathbf{z}_1$ to get an updated position $\mathbf{z}_2 \in \mathbf{Z}$. To find the most similar small sequence $G(\mathbf{z}_\Lambda) \sim \mathbf{x}$, the location of $\mathbf{z} \in \mathbf{Z}$ is optimised in an iterative process via $\lambda = 1, 2, \dots, \Lambda$ backpropagation steps.

We define the loss function $\mathcal{L}$ to map a real time series small sequence $\mathbf{x} \in \mathbf{X}$ to the best latent space location $\mathbf{z} \in \mathbf{Z}$ \cite{Schlegl2017UnsupervisedDiscovery}. The loss function $\mathcal{L}$ has two parts, a residual loss $\mathcal{L}_R$ and a discrimination loss $\mathcal{L}_D$ as follows. 

\noindent \textbf{Residual Loss} $\mathcal{L}_R$ measures the point-wise dissimilarity (e.g. values in timestamps) between real small sequence $\mathbf{x}$ and fake small sequence $G(\mathbf{z}_\lambda)$, and it is defined as
\begin{equation}
    \mathcal{L}_R(\mathbf{z}_\lambda) = \sum \left | \mathbf{x} - G(\mathbf{z}_\lambda) \right |
\end{equation}

\noindent \textbf{Discrimination Loss} A rich intermediate feature representation of the discriminator is used for estimating discriminator loss. That is, the output of an intermediate layer $f(\cdot)$ of the discriminator is used to specify the statistics of an input small sequence. The discriminator loss is defined as
\begin{equation}
    \mathcal{L}_D(\mathbf{z}_\lambda) = \sum \left | f(\mathbf{x}) - f(G(\mathbf{z}_\lambda)) \right |
\end{equation}
The Loss function $\mathcal{L}$ is defined as a weighted sum of residual loss and discrimination loss as follows.
\begin{equation}
    \mathcal{L}(\mathbf{z}_\lambda) = (1-\gamma) \cdot \mathcal{L}_R(\mathbf{z}_\lambda) + \gamma \cdot \mathcal{L}_D(\mathbf{z}_\lambda)
\end{equation}
$\mathcal{L}_R$ enforces the point-wise similarity between the fake small sequence $G(\mathbf{z}_\lambda)$ and the real small sequence $\mathbf{x}$. $\mathcal{L}_D$ enforces the fake small sequence $G(\mathbf{z}_\lambda)$ to lie in the manifold $\mathbf{X}$. This means, both $G$ and $D$ are utilised to update the parameters of $\mathbf{z}$ via backpropagation. In this inverse mapping process, only the parameters of $\mathbf{z}$ are updated via backpropagation; the parameters of $G$ and $D$ are kept fixed.

\begin{figure}[htb!]
    \centering
    \subfloat[Anomaly score $A(\mathbf{x})$ in blue, ground truth $y$ in green and threshold in orange colour]{{\includegraphics[width=5cm]{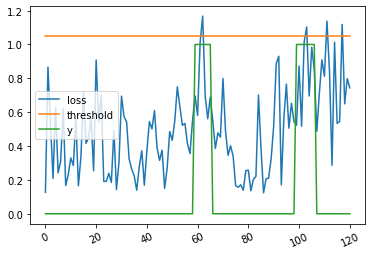}}\label{vis_TAnoGAN_loss}}
    \quad
    \subfloat[System values in blue, the ground truth $y$ in orange and detected anomalies in red colour.]{{\includegraphics[width=5cm]{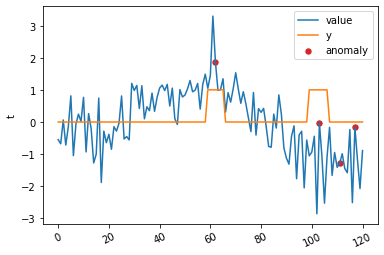}}\label{vis_TAnoGAN_value}}
    \caption{Visualisation of Anomaly Detected by TAnoGan.}%
    \label{fig:vis_TAnoGAN}%
    \vspace{-.6cm}
\end{figure}

\subsection{Detection of Anomalies}
During anomaly detection in a small sequence time series data, we evaluate each small sequence $\mathbf{x}$ as being a normal (i.e. originating from the general data distribution) or anomalous (i.e. deviated from the general data distribution) observation. 

In the adversarial training process, the generator $G$ learns the general data distribution $p_g$ of latent space $\mathbf{Z}$ and the general data distribution $p_{data}$ of real data space $\mathbf{X}$. That is, each fake small sequence $G(\mathbf{z}_\lambda)$ is generated based on the general data distribution of $\mathbf{X}$. In every updating iteration $\lambda$, the loss function $\mathcal{L}$ evaluates the dissimilarity of the fake small sequence $G(\mathbf{z}_\lambda)$ with the real small sequence $\mathbf{x}$. An anomaly score $A(\mathbf{x})$ expressing the fit of a given $\mathbf{x}$ to the general data distribution (i.e. model of normal small sequences) can be directly derived from $\mathcal{L}$. 
\begin{equation}
    A(\mathbf{x}) = (1-\gamma) \cdot \mathcal{R}(\mathbf{x}) + \gamma \cdot \mathcal{D}(\mathbf{x}),
\end{equation}
where the residual score $\mathcal{R}(\mathbf{x})$ and the discrimination score $\mathcal{D}(\mathbf{x})$ are defined by the residual loss $\mathcal{L}_R(\mathbf{z}_\Lambda)$ and the discrimination loss $\mathcal{L}_D(\mathbf{z}_\Lambda)$, respectively, at the final ($\Lambda$th) updating iteration of the mapping procedure to the latent space. The parameter $\gamma$ is a trade-off factor decided empirically. A large anomaly score $A(\mathbf{x})$ indicates an anomalous small sequence whereas a small anomaly score indicates a small sequence fitting to the general data distribution of $\mathbf{X}$ learned by $G$ during adversarial training. 

Figure \ref{fig:vis_TAnoGAN} shows an example of detected anomalies in TAnoGan. The blue graph in Figure \ref{vis_TAnoGAN_loss} shows the losses over the re-constructed small sequences. It shows four peaks over the orange threshold line. Threshold was determined from visual observation and model performance. Referring to the green ground truth line, it indicates that two of the peaks are true positive and other two are false positive. In Figure \ref{vis_TAnoGAN_value}, the blue graph shows the original time sequence of the dataset. The orange ground truth line shows the time spans where anomaly originally happened and the red dots show where TAnoGan identified the anomalies. Referring to the orange ground truth, TAnoGan is able to correctly identify two anomalies. Event though TAnoGan has two false positive, a close observation of the original time series shows that these two points  quite different than the general data distribution.   

\vspace{-2mm}
\section{Empirical Evaluation}
\label{sec:evaluation}
\vspace{-2mm}
The primary objective of experiments is to show the effectiveness of the adversarialy trained generator model, i.e., the proposed TAnoGan model, for identifying anomalies in time series data.  

\vspace{-.2cm}
\subsection{Data Collection}
\label{sec:data_collection}
\vspace{-.1cm}
We use the NAB data collection \cite{Lavin2017ThePaper,Lavin2015EvaluatingBenchmark} in experiments. NAB is a quality collection of time-series data with labelled anomalies \cite{Lavin2015EvaluatingBenchmark}. The current NAB collection contains 58 datasets, each with 1000-22,000 data instances, for a total of 365,551 data points. The datasets come in seven categories. Five of the categories consists of real data and the rest two consists of artificial data. We use all 46 datasets in the five categories of real data. 

Each dataset in NAB is labelled manually using  a documented meticulous procedure. Labelers had to adhere to a set of rules when inspecting datasets for anomalies. A key element of the NAB data collection is the inclusion of real-world datasets with anomalies for which the causes are known. Each row in the NAB time-series datasets contains a time stamp and a single scalar value. Each anomaly label in the datasets is associated with a time span marked by a beginning and ending timestamp. The argument behind anomaly time span is that anomalous data often occurs over time, rather than at a single point. 
We train TAnoGan and baseline models in the unsupervised fashion, i.e. without labels. The ground truths (i.e. NAB provided labels) are only used to evaluate the anomaly detection performances by comparing the predicted anomalies.

For data preparation, we subdivide the original long sequences of time series into smaller time series by taking a sliding window across rows. Each time series small sequence is an observation for models.
Deciding the optimal window length is important in time series study \cite{Li2019MAD-GAN:Networks}. We tried a set of different window sizes to capture the system status at different resolutions, namely $s_w$ = $30 \times i$, $i = 1, 2, ..., 10$. We found that $i=2$ gives reasonably optimal results for models. To capture the relevant dynamics of datasets, the window is applied with shift length of 1 during training. The shift length of window size is used during testing. If a small time series is within (or overlap) the anomaly time span, the smaller time series is considered anomaly \cite{Lavin2017ThePaper,Lavin2015EvaluatingBenchmark}. This subdivision allows for rewarding early detection and penalising considerable late detection as well.

\vspace{-2mm}
\subsection{Evaluation Measures}
\vspace{-2mm}
We used six standard classification evaluation measures: Accuracy (Ac), Precision (Pr), Recall (Re), F$_1$ Score (F$_1$), Cohen Kappa (CK) and Area Under Curve (AUC). A detailed description of these measures are available at \cite{Bashar2020RegularisingSet}. 

\vspace{-2mm}
\subsection{Baseline Models}
\label{sec:baselines}
\vspace{-2mm}
We have implemented eight state-of-the-art anomaly detection models as baseline models to compare the performance of the proposed TAnoGan model. 

\begin{enumerate}[wide, labelwidth=!, labelindent=0pt]

\item \textbf{Multivariate Anomaly Detection for Time Series Data with GAN (MadGan)} \cite{Li2019MAD-GAN:Networks}: It uses an LSTM network with depth 3 and 100 hidden (internal) units for the generator and an LSTM network with depth 1 and 100 hidden units for discriminator.

\item \textbf{Auto Encoder (AutoEn)} \cite{Borghesi2019AnomalySystems,Zhou2017AnomalyAutoencoders}: It has three layers: (a) an LSTM encoder layer with 256 hidden units, (b) an LSTM decoder layer with 512 hidden units, (c) a dense final output layer with number of units same as the small time series length as discussed in Section \ref{sec:data_collection}. Both encoder and decoder have 20\% dropout.

\item \textbf{Vanilla Long Short-Term Memory (VanLstm)} \cite{Malhotra2015LongSeries,hochreiter1997long}:
LSTM learns sequences in the data and makes prediction based on a given sequence. We consider a data point as anomaly when its value predicted by LSTM based on previous sequence is distant from the original value. More specifically, given a window of time series data we predict immediate next value in a long sequence. This model uses two LSTM layers, where the first layer has 256 hidden units and the second layer has 512 hidden units, and each layer uses 20\% random dropout. 
 
\item \textbf{Isolation Forest (IsoF)} \cite{Liu2008IsolationForest}: This traditional model randomly selects a feature and a split value between the maximum and minimum values of the selected feature. A recursive partitioning can be represented by a tree structure. The number of splittings required to isolate a sample is equivalent to the path length from root to the leaf node. This path length, averaged over a forest of such random trees, is a measure of normality. Partitioning produces noticeable shorter paths for anomalies. When the path length for a particular sample is short, it is considered highly likely to be anomaly.

\item \textbf{Gaussian Mixture Model (GMM)} \cite{bishop2006linear,Reynolds2009GaussianModels}: It is one of the most popular Unsupervised Clustering approach, where $K$ Gaussians curves are fitted to the training data with $K$ clusters. Each Gaussian has three parameters: mean $\mu$ that defines the centre of the Gaussian, variance $\sigma$ that defines the width of the Gaussian and a mixing probability $\pi$ that defines the size of the Gaussian. The mixing probability meets the condition, $\sum_1^K \pi_k = 1$. Maximum likelihood is used to fit each Gaussian to the data points belonging to each cluster to determine the optimal values for the parameters. Gaussian Mixture Models allow assigning a probability $p$ to each data point of being generated by one of $K$ Gaussian distributions. An outlier is defined as $p < \tau$ for each cluster, where $\tau$ is the threshold. 

\item \textbf{One Class Support Vector Machine (OCSvm)} \cite{ScholkopfSupportDetection}: This unsupervised technique uses a smallest possible hypersphere to encompass instances. The distance from the hypersphere to an instance located outside of the hypersphere is used to identify if the instance is anomalous or not.   

\item \textbf{Bidirectional LSTM in GAN (BiLstmGan)} \cite{Zhu2019ElectrocardiogramNetwork}:
It uses a bidirectional LSTM network with depth 3 and 50 hidden (internal) units for the generator and a bidirectional LSTM network with depth 1 and 50 hidden units for discriminator.

\item \textbf{CNN in GAN (CnnGan)} \cite{Schlegl2017UnsupervisedDiscovery}: 
It uses four staked CNN for each of generator and discriminator. Each CNN has two convolutional layers with 20\% dropout and ReLU activation and kernel size of 256. The number of units in each covolutional layer is the same as the length of small sequence. The final layer is a fully connected layer.      

\end{enumerate}

\begin{figure*}[htb!]
    \centering

    \subfloat{{\includegraphics[width=5.5cm]{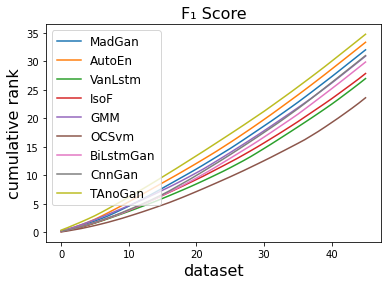}}}
    \quad
    \subfloat{{\includegraphics[width=5.5cm]{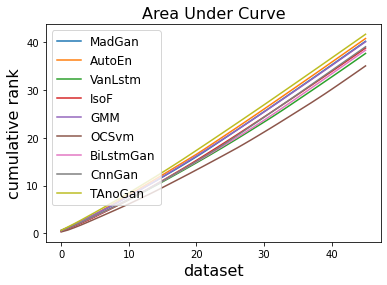}}}
    \quad
    \subfloat{{\includegraphics[width=5.5cm]{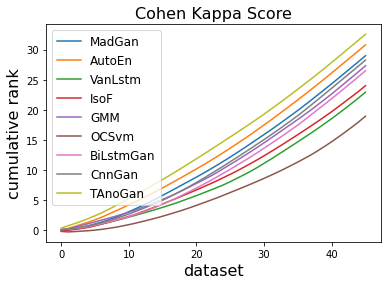}}}
    
    \subfloat{{\includegraphics[width=5.5cm]{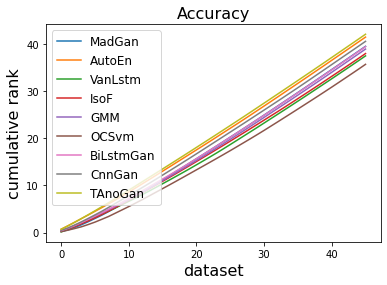}}}
    \quad
    \subfloat{{\includegraphics[width=5.5cm]{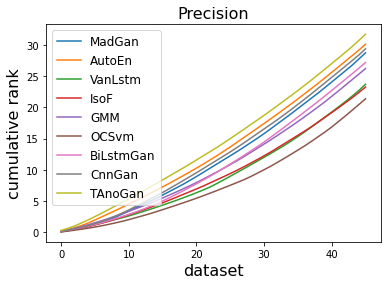}}}
    \quad
    \subfloat{{\includegraphics[width=5.5cm]{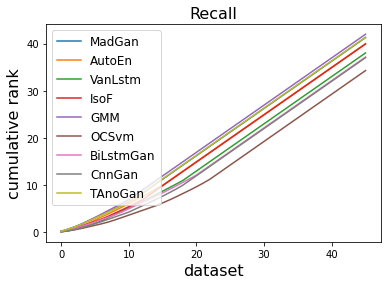}}}
    
    \caption{The cumulative ranking obtained by individually ranking each model per dataset per measure}%
    \label{fig:cr}%
\end{figure*}

\begin{figure*}[htb!]
    \centering
    \subfloat{{\includegraphics[width=3cm]{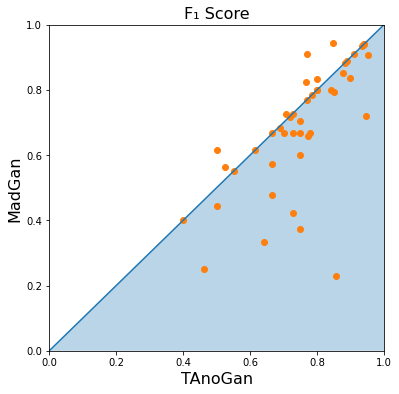}}}
    \quad
    \subfloat{{\includegraphics[width=3cm]{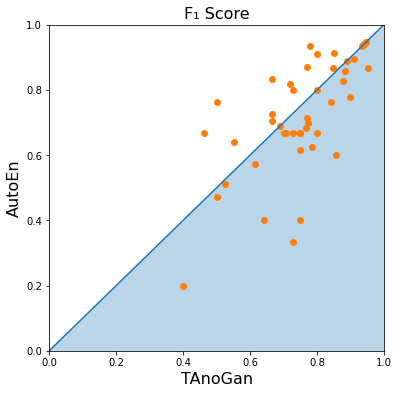}}}
    \quad
    \subfloat{{\includegraphics[width=3cm]{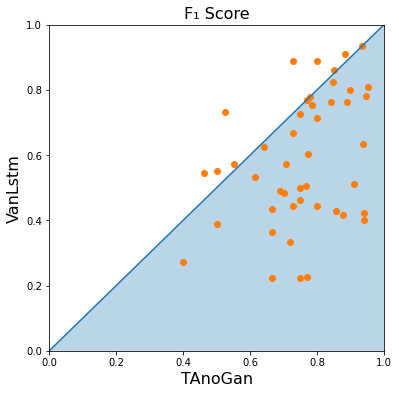}}}
    \quad
    \subfloat{{\includegraphics[width=3cm]{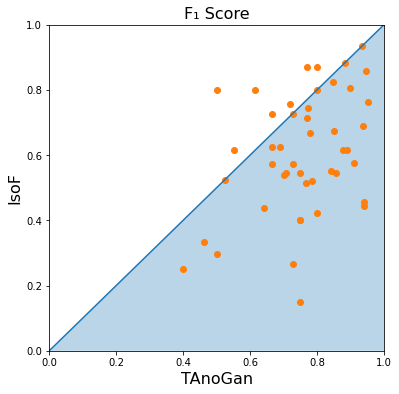}}}
    
    \subfloat{{\includegraphics[width=3cm]{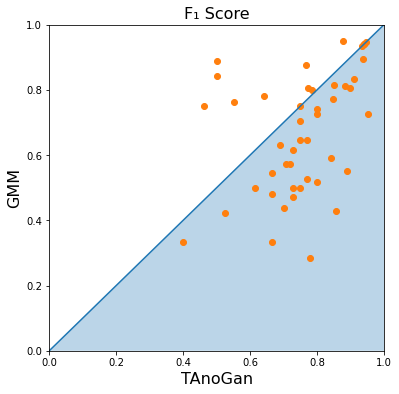}}}
    \quad
    \subfloat{{\includegraphics[width=3cm]{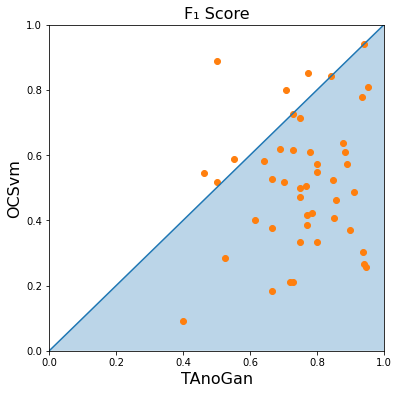}}}
    \quad
    \subfloat{{\includegraphics[width=3cm]{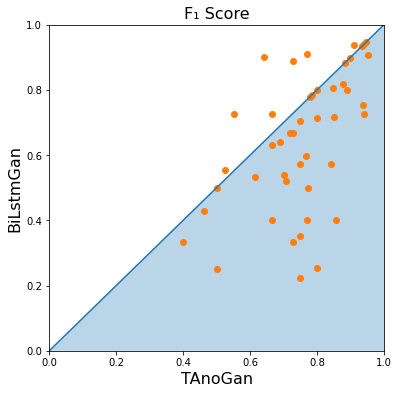}}}
    \quad
    \subfloat{{\includegraphics[width=3cm]{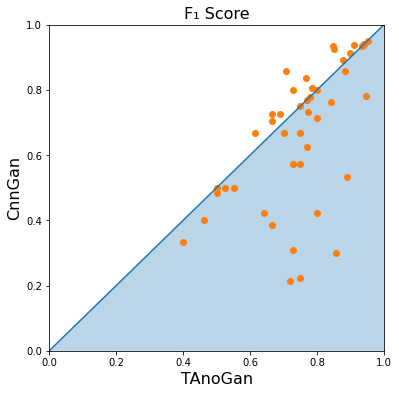}}}

    \caption{Pairwise performance comparison against baseline models. Each dot indicates a dataset. A dot in the blue shade triangle indicates TAnoGan performs better than its competitor and dots in white triangle indicate otherwise, with dots on the diagonal indicating equal performance of both the models.}%
    \label{fig:pwc}%
    \vspace{-.6cm}
\end{figure*}

\begin{figure}[htb!]
    \centering
    \includegraphics[width=6cm]{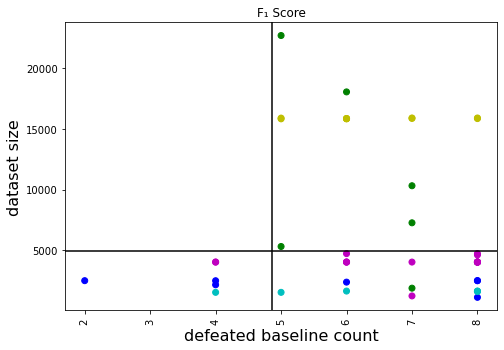}
    \caption{Each point in the scatter plot represents one of the 46 datasets and the colour indicates its membership to one of the five groups (or domain) in NAB collection. The X axis shows the number of baseline models outperformed by TAnoGan. There are a total of eight baseline models.The Y axis shows the size of each dataset. }
    \label{fig:beat_dist}
    \vspace{-.6cm}
\end{figure}

\begin{figure*}[htb!]
    \centering
    
     \subfloat{{\includegraphics[width=5.5cm]{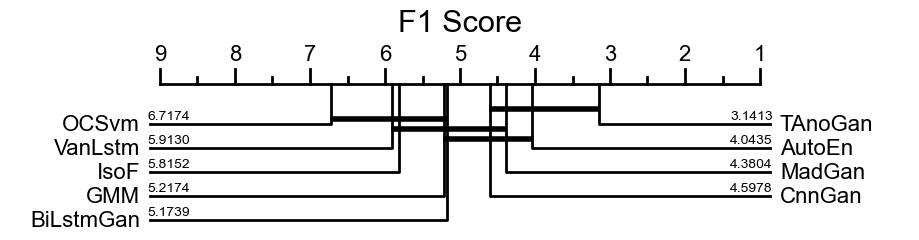}}}
    \quad
    \subfloat{{\includegraphics[width=5.5cm]{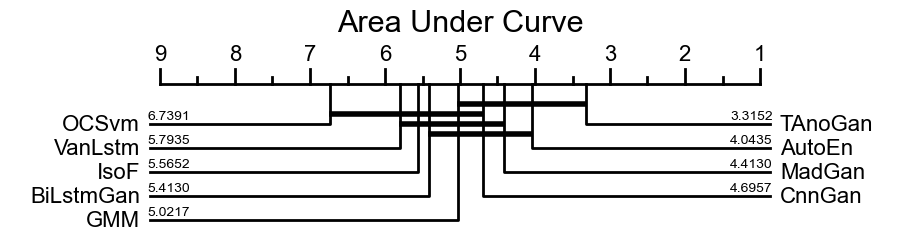}}}
    \quad
    \subfloat{{\includegraphics[width=5.5cm]{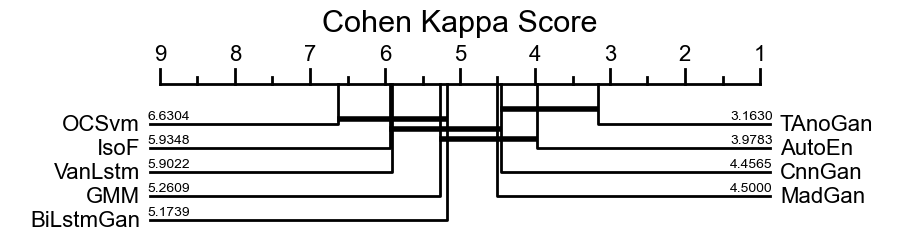}}}
    
    \subfloat{{\includegraphics[width=5.5cm]{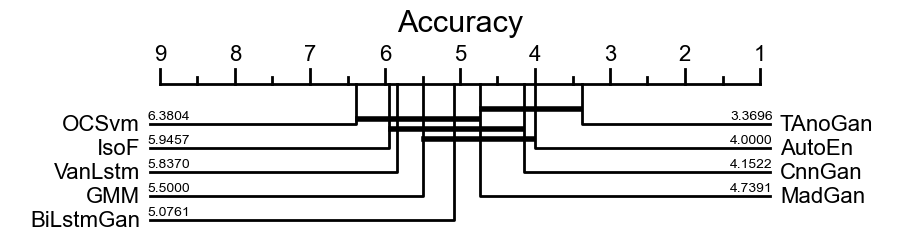}}}
    \quad
    \subfloat{{\includegraphics[width=5.5cm]{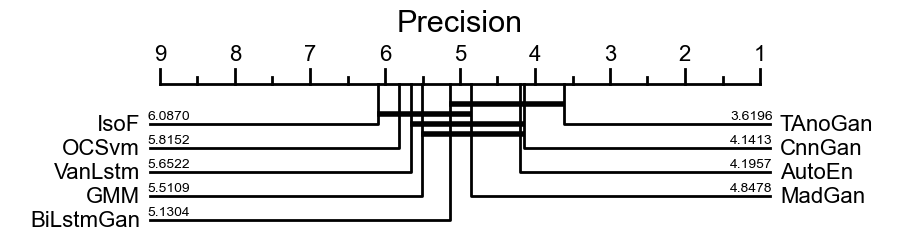}}}
    \quad
    \subfloat{{\includegraphics[width=5.5cm]{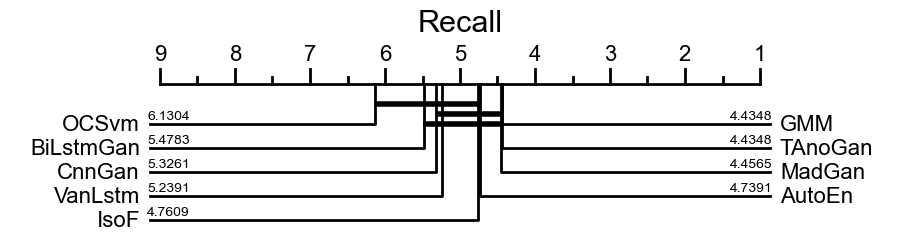}}}
    
    \caption{The critical difference diagram of all the models for different measures}%
    \label{fig:cd}%
    \vspace{-.6cm}
\end{figure*}

\vspace{-2mm}
\subsection{Results}
\vspace{-2mm}
Figure \ref{fig:cr} shows the relative ranking of TAnoGan with baseline models by computing the cumulative ranking for all the models over all the datasets. It is a method-wise accumulation of results measured on a evaluation criterion spanned across all datasets. To compute the cumulative ranking, first we rank each individual method for each dataset and then sum up their ranks according to a criterion. Figure \ref{fig:cr} shows that TAnoGan ranks higher than all baseline models for F$_1$ Score, Cohen Kappa Score, Precision, Accuracy and Area Under Curve. 

Figure \ref{fig:pwc} shows the pairwise comparison of TAnoGan and each baseline model. The number of dots in blue shade triangle indicates the number of datasets where TAnoGan performs better than a baseline model for the most important measure of anomaly detection namely F$_1$ measure. The figure shows that TAnoGan performs better in most datasets as compared with all the baseline modes. 

Figure \ref{fig:beat_dist} shows an analysis of model performance, dataset size and dataset domain (i.e. a category in NAB). This figure reveals that TAnoGan has never performed worse than at least one of the compared algorithms, as the figure includes all of the 46 datasets. The right side of this diagram indicates TAnoGan outperforms majority of the models. Almost all the datasets appear in the right side, which means TAnoGan outperforms majority of the models in almost all the datasets. The upper part of this diagram indicates relatively larger datasets. The diagram shows that when a dataset size is more than 5000 instances, TAnoGan always outperforms majority of the models. 

When a dataset size is less than 5000 instances, TAnoGan outperforms majority of the models in most of the datasets except a few. This means, TAnoGan successfully handles most of the very small datasets but there are still room for improvement for some of the very small datasets. The colour of each dot indicates the domain of the dataset. The same colours are distributed in both left and right sizes of the figure, which means TAnoGan is not sensitive to domain of the datasets.  

Figure \ref{fig:cd} shows the critical difference diagram. It is computed by evaluating each of the methods per dataset using a particular evaluation criterion. \cite{IsmailFawaz2019DeepReview} explained the effectiveness of critical difference in ranking performance of models when the number of datasets for evaluation is high. Critical difference diagram arranges the models on a horizontal line in ascending order of their scores, where lower score means better performance \cite{Demsar2006StatisticalSets}. 

The critical difference diagram shows that TAnoGan has the lowest score, indicating its superior performance over all other models. More specifically, TAnoGan outperforms all baseline methods in all measures except Recall. Recall of TAnoGan is near to GMM and better than other models. Even though GMM has marginally better recall than TAnoGan, GMM has poor $F_1$ scores compared with TAnoGan. This indicates that GMM outputs a lot of false positive, resulting in low precision (the precision diagram also show the same). Better recall is desirable in anomaly detection but the excessive false positives result in a higher cost for investigating many false alarms. A balance in both recall and precision is needed, as indicated by a higher $F_1$ score, which is achieved by TAnoGan. 

\vspace{-2mm}
\subsection{Discussion}
\vspace{-2mm}

All these results justify the effectiveness of GAN for anomaly detection in the time series data. The general distribution of a dataset learned by generative model proves to be helpful in isolating anomalies from the normal ones. 

TAnoGan uses LSTM layers in both generator and discriminator. The performance improvements of TAnoGan over VanLstm \cite{Malhotra2015LongSeries,hochreiter1997long} and AutoEn \cite{Borghesi2019AnomalySystems,Zhou2017AnomalyAutoencoders} models in cumulative ranking, pairwise comparison and critical difference diagram indicate the advantage of adversarial training for learning general distribution of datasets. AutoEn is a well known deep learning-based anomaly detection model. It encodes the time series data to a lower dimension to learn a general distribution of data then decodes the encoded value to a higher dimension to reconstruct the data. This gives AutoEn advantage over VanLstm to learn the general distribution better. Therefore, AutoEn performs significantly better than VanLstm. The generator in TAnoGan is similar to the decoder of AutoEn but the adversarial training allows TAnoGan to learning the general data distribution better than AutoEn.

OCSvm \cite{ScholkopfSupportDetection} provides the poorest performance amongst all models in anomaly detection. A possible reason may be the use of a linear kernel that cannot separate many nonlinear data points. TAnoGan significantly outperforms OCSvm which indicates that TAnoGan can learn nonlinear patterns in the dataset.

IsoF \cite{Liu2008IsolationForest} uses individual features (values in each timestamp of a small sequence in our experiments) in an instance to isolate the instance from the rest. It cannot utilise the nonlinear interaction between these features, while LSTM used in TAnoGan can utilise those nonlinear interaction. As a result, TAnoGan performs better than IsoF. 

GMM \cite{bishop2006linear,Reynolds2009GaussianModels} is flexible in terms of learning co-variance in data distribution, i.e. its co-variance structure is unconstrained. This flexibility allows GMM to learn many complicated data distribution. As a result, GMM achieves significantly better results than other traditional models. However, TAnoGan achieves better results than GMM that indicates TAnoGan can learn complicated data distribution effectively. 

MadGan \cite{Li2019MAD-GAN:Networks}, BiLstmGan \cite{Zhu2019ElectrocardiogramNetwork}, and CnnGan \cite{Schlegl2017UnsupervisedDiscovery} have similar architecture as of TAnoGan. By using progressively higher number of hidden units in generator, TAnoGan performs better than these models that use the same number of hidden units in all the internal layers of generator. The small number of hidden units in earlier layers allows decoding the fine-grained properties of sequences and the larger number of hidden units in later layers allows decoding the coarse-grained properties of sequences. This architecture enables to achieve a hierarchical decoding of sequences that allows to focus on variation of properties in different abstraction levels of the sequences. As small datasets do not have enough instances, learning variation of properties in different abstraction levels can help in generalising the generator.

Overall, TAnoGan consistently outperformed the popular unsupervised anomaly detection models including traditional models and neural network-based models.

\vspace{-4mm}
\section{Conclusion}
\label{sec:conclusions}
\vspace{-2mm}
We proposed a novel method, called TAnoGan, to explore the use of GAN for detecting anomalies in time series when a small number of data points are available. It uses a generator to learn the general data distribution of dataset and an inverse mapping to map sequence to latent space. Sequences mapped to latent spaces are reconstructed by the generator and the reconstruction loss is used to estimate anomaly scores to detect anomalies. By using the progressively higher number of hidden units in the LSTM layers of generator, TAnoGan can effectively detect anomalies in small datasets. 

The experimental results in 46 real-world time series datasets show that TAnoGan performs superior over traditional and neural network models. Determining an optimal window length and model instability are two known issues for GAN-based anomaly detection in time series. GAN-based anomaly detection in time series is sensitive to number of epochs. It will be worthwhile to explore TAnoGan with large architectures in big datasets and compare against more baseline models. We will investigate these issues in future work.




\balance
\vspace{-2mm}
\bibliographystyle{IEEEtran}
\bibliography{references}

\end{document}